\documentclass[10pt,twocolumn,letterpaper]{article}

\usepackage{iccv}
\usepackage{times}
\usepackage{epsfig}
\usepackage{graphicx}
\usepackage{amsmath}
\usepackage{amssymb}
\usepackage{graphicx}
\usepackage{amsmath}
\usepackage{amssymb}
\usepackage{booktabs}
\usepackage{caption}
\usepackage[table,xcdraw]{xcolor}
\usepackage{lipsum}
\usepackage[mathscr]{euscript}
\usepackage{makecell}
\usepackage{multirow}
\usepackage{breqn}
\usepackage{amsmath}

\DeclareSymbolFont{rsfs}{U}{rsfs}{m}{n}
\DeclareSymbolFontAlphabet{\mathscrsfs}{rsfs}

\usepackage[table,xcdraw]{xcolor}

\usepackage[pagebackref=true,breaklinks=true,letterpaper=true,colorlinks,bookmarks=false]{hyperref}

\iccvfinalcopy 

\def\iccvPaperID{15} 

\ificcvfinal\fi

\begin{document}

\title{PRAT: PRofiling Adversarial aTtacks}

\author{Rahul Ambati$^1$ \qquad Naveed Akhtar$^2$ \qquad Ajmal Mian$^2$ \qquad Yogesh S Rawat$^1$ \\ $^1$University of Central Florida \qquad $^2$The University of Western Australia \\
\tt\small rahul.ambati@knights.ucf.edu \qquad \{naveed.akhtar,ajmal.mian\}@uwa.edu.au \qquad yogesh@crcv.ucf.edu
}

\maketitle
\ificcvfinal\thispagestyle{empty}\fi

\begin{abstract}
\vspace{-3mm}
Intrinsic susceptibility of deep learning to adversarial examples has led to a plethora of  attack techniques with a broad common objective of fooling deep models. However, we find slight compositional differences between the algorithms achieving this objective. These differences leave traces that provide important clues for attacker profiling in real-life scenarios. Inspired by this, we introduce a novel problem of \textit{`PRofiling Adversarial aTtacks' (PRAT)}. Given an adversarial example, the objective of PRAT is to identify the attack used to generate it. Under this  perspective, we can systematically group existing attacks into different families, leading to the sub-problem of attack family identification, which we also study. To enable PRAT analysis, we introduce a large \textit{`Adversarial Identification Dataset' (AID)}, comprising over 180k adversarial samples generated with 13 popular attacks for image specific/agnostic white/black box setups. We use AID to devise a novel framework for the PRAT objective. Our framework utilizes a Transformer based Global-LOcal Feature (GLOF) module to extract an approximate signature of the adversarial attack, which in turn is used for the identification of the attack. Using AID and our framework, we provide multiple interesting benchmark results for the PRAT problem. The dataset and the code are available at \href{https://github.com/rahulambati/PRAT}{https://github.com/rahulambati/PRAT}
\end{abstract}

\vspace{-5mm}
\section{Introduction}
Deep learning is currently at the center of many emerging technologies, from autonomous vehicles to numerous security applications. However, it is also well-established that deep networks are  susceptible to adversarial attacks~\cite{akhtar2018threat,chakraborty2018adversarial}. This intriguing weakness of deep learning, which is otherwise known to supersede human intelligence in complex tasks~\cite{silver2017mastering}, has attracted an ever-increasing interest of the research community in the last few years~\cite{chakraborty2021survey}. This has led to a wide range of adversarial attacks that can effectively fool deep learning. Although adversarial attacks have also led to research in defenses, there is a consensus that defenses currently lack efficacy. Many of them are easily broken, or become ineffective by changing the attack strategy~\cite{akhtar2021advances}.

Incidentally, deep learning in practice is still widely open to malicious manipulation through adversarial attacks \cite{chakraborty2018adversarial}. It is yet to be seen if this technology can retain its impressive performance while also demonstrating robustness to adversarial attacks. Until an adversarially robust high-performing deep learning framework is developed, practitioners must account for the adversarial susceptibility of deep learning in all applications. These conditions give rise to an important practical problem of `attacker profiling'. In real-life, understanding the attacker's abilities can allow counter-measures even outside the realm of deep learning. However, the current literature on adversarial attacks on  is almost completely void of any exploration along this line. From the pragmatic viewpoint, the primal question of this potential research is, \textit{``given an adversarial example, which attack algorithm was used to generate it?"}.

\begin{figure}[t!]
    \centering
    \includegraphics[width = \linewidth]{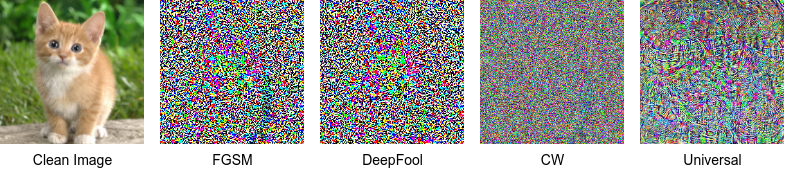}
    \caption{\small Despite their imperceptibility, adversarial perturbations contain peculiar patterns. Perturbations generated using the popular methods FGSM, DeepFool, CW and UAP Attacks are shown.} 
    \label{fig:teaser}
    \vspace{-6mm}
\end{figure}

In this work, we take the first systematic step towards answering this question with PRofiling Adversarial aTtacks (PRAT). Focusing on the \textit{additive adversarial perturbations}, our aim is to explore the extent to which a victim is able to identify its attacker by analyzing only the adversarial input. To explore this new direction, it is imperative to curate a large database of adversarial samples. To that end, we introduce Adversarial Identification Dataset (AID) which consists of over 180k adversarial samples, generated with 13 popular attacks in the literature. AID covers input-specific and input-agnostic attacks and considers white-box and black-box setups. We select these attacks considering the objective of retracing the attacker from the adversarial image. 

We use AID to explore PRAT with a proposed framework that is built on the intuition that attack algorithms leave their peculiar signatures in the adversarial examples. As seen in Fig.~\ref{fig:teaser}, these traces might reveal interesting information that can help in profiling the attacker. Our technique works on the principle of extracting those signatures. At the center of our framework is a signature extractor which is trained to extract input-specific signatures. Unlike random noise, these traces contain global as well as local structure. We design a signature extractor consisting Global-LOcal Feature extractor (GLOF) modules that combine CNN's ability to learn local structure \cite{lecun1995convolutional} and transformer's capability to capture global information \cite{vaswani2017attention,hanruiwang2020hat,dosovitskiy2020vit}. These signatures contain information which corresponds to the attack algorithm and we use this signature to identify the attack leveraged to generate the adversarial example. 

\vspace{1mm}
Our contributions are summarized as follow.
\begin{itemize}
    \vspace{-2mm}
    \item We put forth a new problem of PRofiling Adversarial aTtacks (PRAT), aimed at profiling the attacker. We formalize PRAT to provide a systematic guideline for research in this direction.
    \vspace{-2mm}
    \item We propose an effective framework to provide the first-of-its-kind solution to the PRAT problem which consists of a hybrid Transformer network that combines the capabilities of CNNs and attention networks targeted to solve PRAT.  
    \vspace{-2mm}
    \item We introduce a large Adversarial Identification Dataset (AID), comprising 180k+ adversarial samples generated with 13 different attacks. AID is used to extensively study PRAT, leading to promising results.     
\end{itemize}

\vspace{-2mm}
\section{Related Work}
Adversarial attacks and defenses are currently a highly active research direction. 
Our discussion here focuses on the relevant aspects of this direction with representative existing techniques. 
The discovery of adversarial susceptibility of deep learning was made in the context of visual classifiers~\cite{szegedy2013intriguing}. \cite{szegedy2013intriguing} demonstrated that deep models can be fooled into incorrect prediction by adding imperceptible adversarial perturbations to the input. 
Hence, to efficiently compute adversarial samples (for adversarial training), \cite{goodfellow2014explaining} proposed the Fast Gradient Sign Method (FGSM). Conceptually, the FGSM takes a single gradient ascend step over the loss surface of the model w.r.t.~input to compute the adversarial perturbation. 

\cite{kurakin2016adversarial} enhanced FGSM to iteratively take multiple small steps for gradient ascend, thereby calling their strategy Basic Iterative Method (BIM). A similar underlying scheme is adopted by the Projected Gradient Descent (PGD) attack \cite{madry2017towards}, with an additional step of projecting the gradient signals on a pre-fixed $\ell_p$-ball to constrain the norm of the resulting perturbation signal. All the above attacks must compute model gradient to compute the perturbations. Hence, we can categorise them as gradient-based attacks. Moreover, the gradient computation normally requires complete knowledge of the model itself hence categorized as white-box attacks. 
Other popular gradient based attacks include Carlini \& Wagner attack~\cite{carlini2017towards}, DeepFool~\cite{moosavi2016deepfool} and Jacobian Saliency Map Attack (JSMA)~\cite{papernot2016limitations}.   

Black-box attacks do not assume any knowledge of the model, except its predictions. The most popular streams of black-box attacks are query-based attacks, which allow the attacker to iteratively refine an adversarial example by sending the current version to the remote model as a query. The model's prediction is used as feedback for improving the adversarial nature of the input. If the attacker only receives the model decision (not its confidence score), then such an attack is called a decision-based attack. Currently, the decision based attacks are more popular in black-box setups due to their pragmatic nature. A few recent representative examples in this category include \cite{rahmati2020geoda}, \cite{shi2020polishing}, \cite{du2019query}, \cite{li2020qeba}.  

With the discovery of adversarial samples, there is an increased interest in devising defences, of which, the most popular strategy is adversarial training \cite{goodfellow2014explaining,kannan2018adversarial,madry2017towards,tramer2018ensemble,zhang2019theoretically}. 
  
The existing literature also covers a wide range of other defense techniques, from augmenting the models with external defense modules \cite{Qin2020Detecting,DBLP:conf/eccv/LiZPRSKSC20,deng2021libre} to certified defenses \cite{katz2017reluplex,tjeng2018evaluating,Croce2020Provable}. Here, we emphasize that these defenses generally come at considerable computational cost and degradation in model performance on clean inputs.       

Instead of proposing yet another defense, we take a different perspective on addressing the adversarial susceptibility of deep learning. Assuming a deployed model, we aim at identifying the capabilities of the attacker. Such an attacker profiling can help in adversarial defenses outside the realm of deep learning. This is more practical because it can eventually allow deep learning models to disregard intrinsic/appended defensive modules that result in performance degradation, causing deep learning to lose its advantage over other machine learning frameworks.
\vspace{-1mm}
    
\section{The PRAT Problem}
\vspace{-1mm}
The PRofiling Adversasrial aTtacks (PRAT) problem is generic in nature. However, we limit its scope to visual classifiers in this work for a  systematic first-of-its-kind study. 
Let $\mathcal{C}(.)$ be a deep visual classifier such that $\mathcal{C}({\bf I}): {\bf I} \rightarrow \boldsymbol\ell$, where ${\bf I} \in \mathbb R^m$ is a natural image 
and $\boldsymbol\ell \in \mathbb Z^+$ is the output of the classifier. For   attacking $\mathcal{C}(.)$, an  adversary seeks a signal $\boldsymbol\rho \in \mathbb R^m$ to achieve $\mathcal{C}({\bf I} + \boldsymbol\rho) \rightarrow \tilde{\boldsymbol\ell}$, where $\tilde{\boldsymbol\ell} \neq {\boldsymbol\ell}$. To ensure that the manipulation to a clean image is humanly imperceptible, the perturbation $\boldsymbol\rho$ is norm-bounded, e.g.,~by enforcing $|| \boldsymbol\rho||_p < \eta$, where $||.||_p$ denotes the $\ell_p$-norm of a vector and `$\eta$' is a pre-defined scalar. More concisely, the adversary seeks $\boldsymbol\rho$ that satisfies
\vspace{-1mm}
\begin{align}
    \mathcal{C}({\bf I} + \boldsymbol\rho) \rightarrow \tilde{\boldsymbol\ell}~~\text{s.t.}~ \tilde{\boldsymbol\ell} \neq {\boldsymbol\ell}, ||\boldsymbol\rho||_p < \eta.
    \label{eq:attackEq}
\end{align}
The above formulation underpins the most widely adopted settings for the adversarial attacks, where $\boldsymbol\rho$ is a systematically  computed additive signal. From our PRAT perspective, we see this signal as a function  $\boldsymbol\rho (\mathcal A,  \{{\bf I}\}, \mathcal C)$, where $\mathcal A$ identifies the algorithm used to generate the perturbation and  $\{{\bf I}\}$ indicates that $\boldsymbol\rho$ can be defined over a set of images instead of a single image.

In practice, the targeted model $\mathcal C$ must already be deployed and the input ${\bf I}$ fixed during an attack leaving$\mathcal A$ as the point of interest for the PRAT problem. For clarity, we often refer to $\mathcal A$ directly as `attack' in the text. To abstract away the  algorithmic details, we can  conceptualize $\mathcal A$ as a function $\mathcal A( \{\boldsymbol \varphi\}, \{\boldsymbol \psi \})$, where  $\{\boldsymbol \varphi\}$ denotes a set of abstract design hyper-parameters and  $\{\boldsymbol \psi \}$ is a set of numeric hyper-parameters. 
To exemplify, the choice of the scope of the adversarial objective, e.g.~universal vs image-specific, is governed by an element in $\{\boldsymbol \varphi\}$. Similarly, the choices of `$\eta$' or `$p$' values in Eq.~(\ref{eq:attackEq}) are overseen by the elements of $\{\boldsymbol \psi \}$. Collectively, both sets contain all the hyper-parameters available to an attacker to compute $\boldsymbol{\rho}$.

We are particularly interested in the design choices made under $\{\boldsymbol \varphi\}$.
In the considered settings, $\{\boldsymbol \varphi\}$ is a finite set because each of its elements, i.e., $\varphi_i \in \{\boldsymbol \varphi\}$, governs a choice along a specific design dimension under the practical constraint that the attack must achieve its fooling objective. Nevertheless, in this work, we are not after exhaustively listing the elements of $\{\boldsymbol \varphi\}$. Instead, we specify only three representative elements to demonstrate the possibility of attack profiling. These three elements are 1) $\varphi_1$-\textit{model gradient information}, 2) $\varphi_2$-\textit{black-box prediction score information}, and 3) $\varphi_3$-\textit{attack fooling scope}.


It is possible to easily extend the above list to incorporate further design choices. The criterion for a parameter to be enrolled in $\{\boldsymbol\varphi \}$ is that a single choice should cover a range of existing attacks. For instance, $\varphi_1$ can either be \texttt{true} for a family of attacks $\mathcal F_1^a$ of gradient-based attacks and can be \texttt{false} for non-gradient based attack family $\mathcal F_1^b$. Similarly, when $\varphi_2 = \texttt{true}$, we get an attack family $\mathcal F_2^a$ of score-based black-box attacks\cite{ma2021simulator,Huang2020Black-Box}, and $\varphi_2 = \texttt{false}$ yields $\mathcal F_2^b$ that represents decision-based attacks\cite{yang,ACFH2020square,Chen2020BoostingDB}. Similarly, $\varphi_3 = \texttt{true}$ results in the $\mathcal F_3^a$ representing universal attacks and $\varphi_3 = \texttt{false}$ corresponds to the family $\mathcal F_3^b$ consisting input-specific attacks.

In the above formalism, $\mathcal F_i^x  \cap \mathcal F_i^y = \emptyset$ always holds for the resulting attack families. However, we must allow $\mathcal F_i^x  \cap \mathcal F_j^x \neq  \emptyset$ because an attack family resulting from $\varphi_i$ may still make choices for $\varphi_{j \neq i}$ without any constraint.
Let $\mathcal F_i = \{f_1^i, f_2^i,..., f_Z^i\}$ denote the $i^{\text{th}}$ attack family with `$Z$' adversarial attacks that are formed under $\varphi_i$ such that all $f_z^i \in \mathcal F_i$ satisfy the constraint in Eq.~(\ref{eq:attackEq}). Then, $f_z^i({\bf I}) \rightarrow \tilde{\bf I}$ s.t.~$\mathcal C (\tilde{\bf I}) \rightarrow \tilde{\boldsymbol\ell} \neq {\boldsymbol\ell}, ||\boldsymbol\rho||_p < \eta.$ In this setting, the core PRAT problem is a reverse mapping problem that computes $\Psi(\tilde{\bf I}) \rightarrow f_z^i$, given a set of `$N$' attack families $\mathcal F = \{\mathcal F_1, \mathcal F_2,...,\mathcal F_N \}$. We must seek $\Psi(.)$ to solve this.

\section{Adversarial Identification Dataset (AID)}
\vspace{-1mm}
To investigate the PRAT problem, we develop Adversarial Identification Dataset (AID). 
Below, we detail different attacks $\mathcal A$, attack families $\mathcal F$ and their  design and numeric hyper-parameters $(\{\boldsymbol \varphi\}, \{\boldsymbol \psi \})$ considered in AID. 

Most of the existing literature on adversarial attacks concentrates on devising novel attack schemes or robustifying models against the attacks. Multiple existing adversarial attack libraries are available to generate adversarial samples on-the-fly. However, for our problem, it is imperative that we store the generated adversarial perturbations to analyze them for reverse engineering. This motivates the curation of Adversarial Identification Dataset (AID) that comprises perturbations generated by leveraging different attack strategies over a set of images targeting different pre-trained classifiers. In line with our PRAT problem, AID consists of 3 different attack families (\textit{gradient-based}, \textit{decision-based}, and \textit{universal}) with 13 different attack strategies resulting in over 180k samples. We discuss these families next. 

\vspace{-1mm}
\subsection{Attack Families}
\vspace{-1mm}
\textbf{Gradient based attacks:} Gradient based attacks are able to exploit the gradients of the target model to perturb input images. Since the attacker needs access to the gradients, these attacks are typically white box in nature. Our gradient-based attack family consists of \textit{Fast Gradient Sign Method (FGSM)}  \cite{goodfellow2014explaining}, \textit{Basic Iterative Method} \cite{kurakin2016adversarial}, \textit{NewtonFool} \cite{newton}, \textit{Projected Gradient Descent(PGD)} \cite{madry2017towards}, \textit{DeepFool} \cite{deep}, \textit{Carlini Wagner (CW)} \cite{carlini} attacks.

\textbf{Decision based attacks:} Decision-based attacks are  applied in black-box setups where the attacker only has access to the decision of the target model. 
The attacker repeatedly queries the target model and utilizes the decision of the model to curate the perturbation. We consider \textit{Additive Gaussian Noise}, \textit{Gaussian Blur}, \textit{Salt \& Pepper Noise}, \textit{Contrast Reduction}, and  \textit{Boundary Attack} \cite{brendel2018decisionbased} for this family. 

\textbf{Universal attacks:} Universal attacks generalize across a dataset. A single perturbation is sufficient to fool the network across multiple images with a desired fooling probability. Most common approaches to generate universal perturbations either iteratively compute perturbations by gradually computing and projecting model gradients over input batches, or use generative modeling to compute image agnostic perturbations. We consider \textit{Universal Adversarial Perturbation (UAP)}  \cite{Moosavi-Dezfooli_2017_CVPR}, \textit{Universal Adversarial Network (UAN)} \cite{universal} for the universal attack family.

\subsection{Dataset creation}
\vspace{-1.5mm}
\textbf{Benign samples:}
We require clean images to create an adversarial perturbation. 
We utilize ImageNet2012 \cite{ILSVRC15} validation set consisting of 50k images spanning across 1000 classes. We split the validation set into two exclusive parts, 
forming training and test partitions of AID. The training set of perturbed images for AID is generated by randomly choosing 4k images per network per attack from the training partition. Similarly, the test set of perturbed images is generated by randomly choosing 800 images per network per attack from the test partition. Note that each attack image can be computed with different networks i.e. target models. We discuss these in the following section.

{\bf Target models:}
We consider three  target models; ResNet50 \cite{resnet}, DenseNet121 \cite{dense} and InceptionV3 \cite{incept}. Using multiple models ensures that the adversarial samples are not model specific.

{\bf Attack settings:}
In practice, there can be variations in perturbations norm for an attack - a hyper-parameter from $\{ \boldsymbol\psi \}$.  
This variation is incorporated in AID by sampling $\eta$ from a range of values. For attacks constructed under $l_\infty$ norm, we consider a range of $\{1,16\}$ and $\{1,10\}$ for $l_2$ norm based attacks. The procedure of generating the entire dataset as well as the summary statistics are further detailed in the supplementary material of the paper.
We also summarise the considered attacks, their families, and used perturbation norm-bounds in Table~\ref{tab:attacks1}.

\begin{figure*}[t]
  \centering
   \includegraphics[width=1\linewidth]{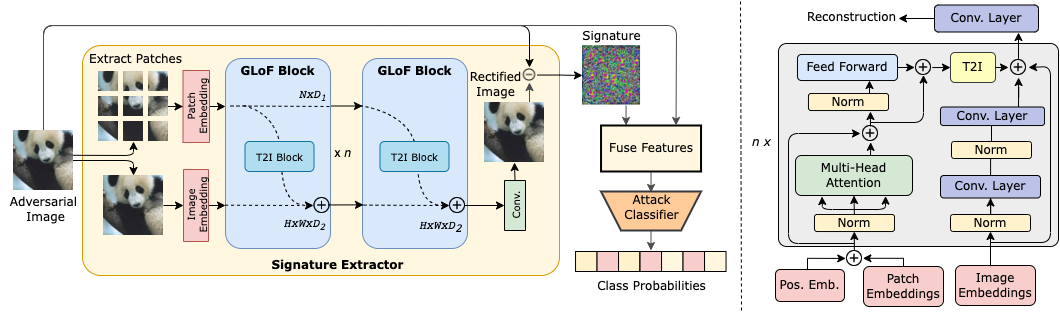}

  \caption{(\textbf{Left}) Schematics of the proposed approach. (\textbf{Right}) GLoF module architecture. In our method, an input adversarial image is passed through a series of GLoF modules. Each GLoF module has two arms; one captures global dependencies, the other captures local features. Extracted signature is fused with the adversarial image and fed to attack classifier.
  }
  \label{fig:pipeline}
  \vspace{-4mm}
\end{figure*}

\vspace{-1mm}
\section{Proposed Approach}
\vspace{-1mm}
Here, we discuss the design choices we consider for solving the PRAT problem $\Psi(\tilde{\bf I}) \rightarrow f_z^i$. A simple approach to solve PRAT could be to build a classifier $C(\tilde{\bf I}) \rightarrow f_z^i$ that identifies the attack leveraged to generate the adversarial input $\tilde{\bf I}$. In such a scenario, the underlying patterns in the perturbation $\rho$ are closely intertwined with the benign sample $\bf I$, thus making the problem much harder. To solve it, we design a signature extractor $\Omega(\tilde{\bf I}) \rightarrow \tilde{\boldsymbol\rho}$ that generates a signature $\tilde{\boldsymbol\rho}$ from the adversarial input s.t.~it lies close to the original perturbation $\boldsymbol\rho$ while preserving patterns helpful in identifying the attacker. The objective of the signature extractor is 
\begin{align}
    \Omega(\tilde{\bf I}) \rightarrow \tilde{\boldsymbol\rho},~~||\tilde{\boldsymbol\rho} - \boldsymbol\rho||_2 = \boldsymbol\delta,~~min(\boldsymbol\delta).
    \label{eq:sig_eq}
\end{align}
While the objective draws similarities with existing problems like denoising/deraining, signature extraction is relatively complex. Noise/rain pertaining to these tasks are largely localized in nature and are visually perceptible in most cases which is not the case for PRAT that makes the problem more challenging and requires methods beyond standard techniques aimed at denoising and other low-level computer vision tasks.\par
Extracted signature is utilized to train a classifier $C$ that identifies the attack. The objective of the classifier is 
\begin{equation}
    C(\tilde{\boldsymbol\rho}) \rightarrow f_z^i,~~where~~f_z^i({\bf I}) \rightarrow \tilde{\bf I},
    \label{eq:clsfr}
\end{equation}
where, $\tilde{\boldsymbol\rho}$ is the generated signature, $f_z^i$ is the $z^{\text{th}}$ attack from the $i^{\text{th}}$ toolchain family. Figure \ref{fig:pipeline} shows an overview of the proposed approach highlighting the signature extractor and the attack classifier.

\begin{table}
  \centering
 \small
  \setlength{\tabcolsep}{5pt}
  \begin{tabular}{c|c|c|c|c}    
    \toprule
    \textbf{label} & \textbf{Attack Method} & \textbf{Family} & \textbf{Setup} & \textbf{NB} \\
    \midrule
    \hline
    0 & PGD \cite{madry2017towards} & Grad. & WB & \textit{$l_{\infty}$} \\
    \hline
    1 & BIM \cite{kurakin2016adversarial} & Grad. & WB & \textit{$l_{\infty}$} \\
    \hline
    2 & FGSM \cite{goodfellow2014explaining} & Grad. & WB & \textit{$l_{\infty}$} \\
    \hline
    3 & DeepFool \cite{deep} & Grad. & WB & \textit{$l_{\infty}$} \\
    \hline
    4 & NewtonFool \cite{newton} & Grad. & WB & \textit{$l_{2}$} \\
    \hline
    5 & CW \cite{carlini} & Grad. & WB & \textit{$l_{2}$} \\
    \hline
    6 & Additive Gaussian \cite{rauber2017foolbox} & Grad. & BB & \textit{$l_{2}$} \\
    \hline
    7 & Gaussian Blur \cite{rauber2017foolbox} & Grad. & BB & \textit{$l_{\infty}$} \\
    \hline
    8 & Salt\&Pepper \cite{rauber2017foolbox} & Grad. & BB & \textit{$l_{\infty}$} \\
    \hline
    9 & Contrast Reduction \cite{rauber2017foolbox} & Dec. & BB & \textit{$l_{\infty}$} \\
    \hline
    10 & Boundary \cite{brendel2018decisionbased} & Dec.  & BB & \textit{$l_{2}$} \\
    \hline
    11 & UAN \cite{universal}& Uni. & WB & \textit{$l_{\infty}$} \\
    \hline
    12 & UAP \cite{Moosavi-Dezfooli_2017_CVPR} & Uni. & WB & \textit{$l_{\infty}$} \\
    \bottomrule
  \end{tabular}
  \caption{\textbf{Summary of the attacks in AID}. Grad., Dec. and Uni.~denote Gradient-based, Decision-based and Universal attacks. BB and WB  denote Black- and White-box attacks. NB is the norm bound on perturbation.
  }
  \label{tab:attacks1}
  \vspace{-4mm}
\end{table}

{\bf Signature Extractor:}
It serves the purpose of extracting a signature with patterns specific to the attack. As shown in Fig.\ref{fig:pipeline}, the signature extractor has two streams of information flow progressing through a series of GLOF modules. Each stream is designed to capture local or global features along with feature sharing across them. GLOF module utilizes convolutional layers to extract local features while attention mechanism applied over image patches help in attaining global connectivity. Conjunction of global and local features help reconstruct a rectified image that lies in the neighborhood of the clean image. Subtracting the rectified image from the adversarial image yields the signature. 

The input adversarial image $\tilde{\boldsymbol{I}}\in \mathbb{R}^{H\times W\times 3}$ (\textit{H}, \textit{W} correspond to image height and width and 3 corresponds to the RGB channels) is split into a series of patches. The patches are flattened and projected onto the embedding space of dimension $\boldsymbol{D_1}$. 
Similar to \cite{dosovitskiy2020vit}, we add positional embeddings to the patch embeddings. The resulting patch embedding is termed $\boldsymbol{T_0}\in \mathbb{R}^{N\times D_1}$ (0 referring to the initial feature level and $N$ referring to the number of patches). 

Alongside, the input image is projected to an embedding dimension $\boldsymbol{D_2}$, by applying a $3\times3$ Conv with $D_2$ features. We term these features $\boldsymbol{Z_0} \in \mathbb{R}^{H\times W\times D_2}$ (0 refers to the initial feature level). Features extracted from previous level ($l-1$) are passed on to the next GLOF module.
\begin{equation}
    \boldsymbol{T_l},\boldsymbol{Z_l} = GLOF(\boldsymbol{T_{l-1}},\boldsymbol{Z_{l-1}});  \hspace{1em} l=1...L 
  \label{eq:glof_level}
\end{equation}
Where $L$ is the number of GLOF modules. The output of the final GLOF module corresponding to the convolutional arm $\boldsymbol{Z_l}$ is transformed to RGB space by applying a $3\times3$ Conv with 3 feature maps resulting in the rectified image $\boldsymbol{I_r} \in \mathbb{R}^{H\times W\times 3}$. Finally, to extract the signature from the rectified image, difference of the rectified and the original image is considered $\tilde{\boldsymbol\rho} = \tilde{\boldsymbol{I}}-\boldsymbol{I_r}$.

{\bf GLOF Module:}
Standard convolutional layers are  good at extracting local patterns \cite{krizhevsky2012imagenet}. On the other hand, transformers are known to be extremely powerful in learning non-local connectivity \cite{Devlin2019BERTPO}. 
As seen in \cite{dosovitskiy2020vit}, vision transformers fail to utilize the local information \cite{li2021localvit}. 
Overcoming these limitations, we propose Global-LOcal Feature extractor (GLOF) module to combine CNN's ability to extract low-level localized features and vision transformer's ability to extract global connectivity across long range tokens. Detailed schematic of the GLOF module is given in Fig.\ref{fig:pipeline}.

The GLOF module at any level receives the local and global features from the previous level.

{\textit{Local features:}}
Embedded 2D image features from the previous layer $\boldsymbol{Z_{l-1}}$ are fed to a ResNet block\cite{resnet} with convolutional, batch norm and activation layers.

{\textit{Global features:}}
Embedded tokens are fed to attention mechanism\cite{vaswani2017attention}. Series of tokens from previous layer $\boldsymbol{T_{l-1}}$ are passed through a multi-head attention layer which calculates the weighted sum. A feed forward network is applied over the attention output consisting of two dense layers that are applied to individual tokens separately with GELU applied over the output of the first dense layer\cite{dosovitskiy2020vit}.

{\textit{T2I Block:}}
Features from the attention arm corresponding to the global connectivity are merged with the convolutional arm. \textbf{Token to Image (T2I)} is responsible for rearranging the series of tokens to form a 2D grid. This transformed grid is passed to a series of convolutional layers to obtain the feature map with the desired depth and is merged with the features from the convolution arm of the GLOF module. The merged features as well as the learned token embeddings are passed to consecutive GLoF modules. 

{\bf Attack Classifier:}
The generated signature is specific to the input. Since the input contains contextual information, we complement the extracted signature with the adversarial input and feed it to the attack classifier. The fusion is done by applying a series of convolutional layers over the signature and the input  separately and concatenating them. 

{\bf Training objective:}
We utilize two  learning objectives in our framework. We use $L_2$ loss to minimize the distance of generated signature $\tilde{\boldsymbol\rho}$ to the raw perturbation $\boldsymbol\rho$. Alongside, the attack classifier is modelled with cross-entropy loss to generate probability scores over a set of classes. 
\vspace{-2mm}

\section{Experiments}
\vspace{-2mm}
We evaluate the performance of the proposed approach on AID under various settings and also present extensive ablations that support the design choices.

{\bf Implementation details:}
The signature extractor comprises of 5 GLOF modules with the attention arm embedding dimension of 768 and the convolutional arm embedding dimension of 64. The T2I block consists of two convolutional layers with kernel size 5 each followed by batch normalization. We use a patch size of 16x16 and 12 attention heads. Each convolutional arm in the GLOF module consists of a ResNet block with 2 convolutional layers of kernel size 5, batch norm and a skip connection. We use DenseNet121\cite{dense} as the attack classifier. Final layers of the attack classifier are adjusted to compute probabilities over 13 classes for attack identification and 3 classes for attack family identification.

{\bf{GLOF Variants:}}
Standard GLOF module consists of convolution and attention arms. We introduce variants of GLOF that exclusively contain either of the arms allowing us to study the contribution of local and global features separately. We term GLOF-C, referring to the GLOF module with only the convolutional arm and GLOF-A, referring to the GLOF module containing only the attention arm.

{\bf Experimental Setup:}
We employ a two-stage training strategy to train the overall pipeline. In the first stage, the signature extractor is trained to produce the rectified image. Benign samples corresponding to the adversarial inputs are used as the ground truth. Adam optimizer and $L_2$ loss are used to pre-train the signature extractor. In the second stage, the overall pipeline with the pre-trained signature extractor is further trained.
We use cross-entropy loss to train the network with Adam optimizer with a learning rate of $1e^{-4}$ and momentum rates of 0.9 and 0.999. We use exponential decay strategy to decrease the learning rate by $5\%$ every 1k iterations. All experiments are conducted on NVIDIA V100 GPU with a batch size of 16. Two stage training helps in faster convergence of the overall network, allows the signature extractor to learn better, and removes the need to retrain it if novel attacks are included.

\begin{table}[t!]
  \centering
  \small
\setlength{\tabcolsep}{0.3em}
  \begin{tabular}{l|c|c|c}    
    \toprule
    \textbf{Method} & \makecell{\textbf{Attack}\\\textbf{Identification}} & \makecell{\textbf{Attack Family}\\\textbf{Identification}} & \makecell{\textbf{no. of}\\\textbf{params}} \\
    \midrule
    \hline
    ResNet50\cite{resnet} &  68.27\% & 80.11\% & 24.7M \\
    ResNet101\cite{resnet} & 71.03\% & 80.38\% & 43.8M \\
    ResNet152\cite{resnet} & 67.03\%  & 78.48\% & 59.5M \\
    DenseNet121\cite{dense} & 73.20\% & 84.21\% & 8.2M \\
    DenseNet169\cite{dense} &  72.22\% & 84.10\% & 14.3M \\
    DenseNet201\cite{dense} &  73.07\% & 81.69\% & 20.2M  \\
    InceptionV3\cite{incept} & 69.96\% & 81.91\% & 22.9M \\
    ViT-B/16\cite{dosovitskiy2020vit} & 63.91\% & 75.89\% & 85.8M\\
    ViT-B/32\cite{dosovitskiy2020vit} & 54.61\% & 72.34\% & 87.4M \\
    ViT-L/16\cite{dosovitskiy2020vit} & 67.28\% & 78.25\% & 303M\\
    ViT-L/32\cite{dosovitskiy2020vit} & 55.23\% & 72.62\% & 305M\\
    \hline
    \makecell{\textbf{Ours}} & \textbf{80.14\%} & \textbf{84.72\%} & 47.8M \\
    \bottomrule
  \end{tabular}
  \caption{\textbf{Performance of different methods on AID} focusing on identifying 13 different attacks and 3 attack families.
  }
  \label{tab:identifying_attacks}
  \vspace{-4mm}
\end{table}

{\bf Evaluation metrics:}
Since the main objective of the PRAT is classification, we use accuracy to compare across several techniques. We also evaluate the performance of the signature extractor using PSNR and SSIM scores calculated over the rectified image and the benign sample.      


{\bf Baselines:}
Since the PRAT problem is first-of-its-kind, we develop several baselines and compare our technique against them. PRAT at its core is a classification problem, we look at the existing visual classifier models and train them accordingly for the PRAT problem. We consider variants of ResNet \cite{resnet}, DenseNet \cite{dense}, Inception \cite{incept} and different versions of Vision Transformer\cite{dosovitskiy2020vit}-\{ViT-B, ViT-L\}as baselines. In line with the original work, ViT-B refers to the Base version of ViT with 12 encoder layers and ViT-L is the Large version with 24 encoder layers. We analyze patch sizes of 16x16 and 32x32 for both the variants.

\vspace{-1mm}
\subsection{Results}
\vspace{-1mm}
{\bf Attack Identification:}
Table \ref{tab:identifying_attacks} reports the results on PRAT problem evaluated on AID under two settings: identifying the attack as well as the attack family. Our approach with the pre-trained signature extractor, feature fusion and the attack classifier achieves \textbf{80.14\%} accuracy on the attack identification and \textbf{84.72\%} on attack family identification. \par

{\bf Comparison with baselines:}
Table \ref{tab:identifying_attacks} compares the performance of our network against other baselines. The top performing compared method, DenseNet121\cite{dense}, is surpassed by our technique in both categories by a margin of 6.94\% in attack identification and 0.51\% in attack family identification. In general, variants of ResNet\cite{resnet} and Inception\cite{incept} under perform when compared with DenseNet\cite{dense} versions. Comparing with versions of ViT\cite{dosovitskiy2020vit}, CNNs have fewer number of parameters and perform much better in both the settings. One reason for this being that ViT requires large amounts of training data. We also observe a drop in accuracy with increase in a patch size from 16x16 to 32x32 suggesting that ViT\cite{dosovitskiy2020vit} struggles to accurately capture the local intrinsic properties as the patch becomes bigger. We also evaluate the performance of Wiener filtering combined with a classifier. This setting achieves 67.55\% compared to 80.14\% by the GLOF based model. It is evident that identifying attack family is simpler compared to identifying the specific attack.

\begin{figure}[t]
  \centering
         \includegraphics[width=0.95\linewidth]{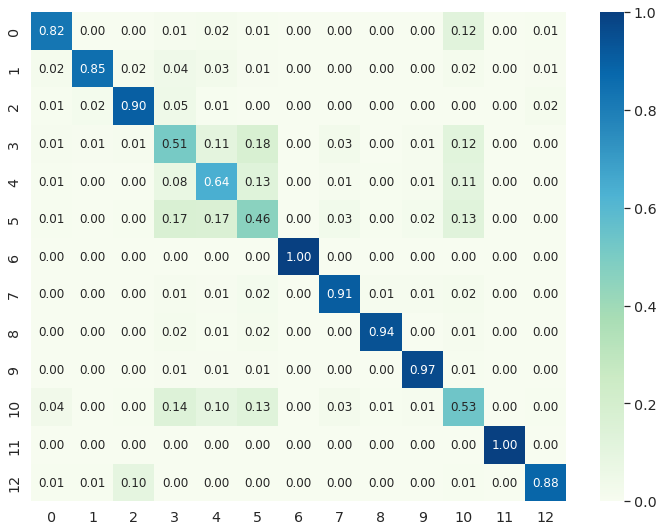}
\caption{\textbf{Confusion matrix:} The labels of the classes are in accordance with the order in Table \ref{tab:attacks1}.}
\label{fig:conf}
\vspace{-4mm}
\end{figure}

\vspace{-1mm}
\subsection{Ablations}
\vspace{-2mm}
Table \ref{tab:ablation} presents the ablation study on the proposed network. Full model refers to the complete pipeline with pre-trained signature extractor and a classifier accepting fused features form the signature and the input which yields 80.14\% on AID. \par
{\bf Effect of pre-training:}
Transformers are known to work well with pre-training. Without pre-training of the signature extractor the accuracy drop to 79.20\%.  \par
{\bf Effect of GLOF module:}
Signature extractor with exclusively GLOF-C variant
yields an accuracy of 78.66\% while its counter part with GLOF-A variant(without CNN blocks) only achieves 73.61\% indicating the importance of both the components for a better performance on PRAT. \par
{\bf Effect of Fusion:}
The fusion module combines the features from the extracted signature and the adversarial image. Removing such fusion module and only relying on the extracted signature results in an accuracy of 78.87\%. \par
{\bf Effect of Signature Extractor:}
While Signature Extractor acts as the crux of the overall pipeline, removing it is no different than the baseline DenseNet121 \cite{deng2021libre} from table \ref{tab:identifying_attacks} which yields 73.20\%.  

\begin{table}[t]
  \centering
   \small
  \begin{tabular}{l|c}    
    \toprule
    \textbf{Method} & \makecell{\textbf{Accuracy}} \\
    \midrule
    \hline
    \makecell{Full model} & \textbf{80.14\%} \\
    \hline
    without pre-training & 79.20\% \\
    without global connectivity- GLoF-C &  78.66\%  \\
    without local connectivity- GLof-A &  73.61\%  \\
    without Feature Fusion & 78.87\%  \\
    without Signature Extractor & 73.20\% \\
    \bottomrule
  \end{tabular}
  \caption{\textbf{Ablation study for Attack Identification.} Full model has a pre-trained signature extractor and a classifier accepting fused product of the signature and   features.}
  \label{tab:ablation}
\end{table}

\begin{table}[t]
\small
        \centering
        \begin{tabular}{l|c|c}    
        \toprule
        \textbf{GLoF Variant} & \makecell{\textbf{PSNR}} & \makecell{\textbf{SSIM}} \\
        \midrule
        \hline
        GLoF-C & 31.49 & 0.88 \\
        \hline
        GLoF-A & 31.53 & 0.87 \\
        \hline
        GloF(Attn.~heads$=4$) & 30.96 & 0.87 \\
        \hline
        GloF(Attn.~heads$=8$) & 30.93 & 0.88 \\
        \hline
        GloF(Attn.~heads$=12$) & \textbf{31.55} & \textbf{0.89} \\
        \hline
        GloF(Attn.~heads$=16$) & 31.54 & \textbf{0.89} \\
        \bottomrule
          \end{tabular}
      \caption{\textbf{Quantitative results of Signature Extractor}. GLoF-C and GLoF-A refer to the variants of GLoF exclusively containing local and global connectivity respectively. }
      \label{tab:abl1}
\end{table}

\begin{table}[t]
\setlength{\tabcolsep}{0.2em}
\small
    \centering
    \begin{tabular}{l|c|c|c|c|c }
    \toprule
       \textbf{\# GLoF}  &  \makecell{$n=1$} & \makecell{$n=3$} & \makecell{$n=5$} & \makecell{$n=7$} & \makecell{$n=9$}\\ \hline
        \textbf{Accuracy} & 79.20\% & 79.65\% & \textbf{80.14\%} & 79.22\% & 79.90\% \\
        \bottomrule
    \end{tabular}
    \caption{Effect of number of GLoF modules $n$ on the performance of attack identification}
    \label{tab:abl2}
    \vspace{-2mm}
\end{table}

\begin{table}
  \centering
  \small
  \begin{tabular}{l|c|c|c|c}    
    \toprule

    \multirow{2}{5em}{\makecell{\textbf{Method}}} & \multirow{2}{3em}{\makecell{\textbf{Train}\\ \textbf{Set}}} & \multicolumn{3}{c}{\makecell{\textbf{Performance on} \\ \textbf{different test sets}}}\\\cline{3-5}
        &  & \textbf{AID-R} & \textbf{AID-D} & \textbf{AID-I} \\ 
    \midrule
    \hline
    \multirow{3}{5em}{\makecell{ResNet50\\\cite{resnet}}} & AID-R & 71.46\% & 65.74\% & 62.90\% \\
        &   AID-D & 66.15\% & 66.88\% & 61.46\%\\
        &   AID-I & 59.69\% & 65.22\% & 66.96\% \\
    \hline
    \multirow{3}{5em}{\makecell{DenNet121\\\cite{dense}}} & AID-R & 70.01\% & 66.89\% & 58.46\% \\
        &   AID-D & 55.77\% & 73.71\% & 53.83\% \\
        &   AID-I & 63.3\% & 66.96\% & 69.54\% \\
    \hline
    \multirow{3}{5em}{\makecell{InceptionV3\\\cite{incept}}} & AID-R & 66.35\% & 60.51\% & 61.29\% \\
        &   AID-D & 63.02\% & 66.05\% & 62.54\% \\
        &   AID-I & 59.21\% & 60.03\% & 68.72\% \\
    \hline
    \multirow{3}{5em}{\makecell{\textbf{Proposed}\\\textbf{Approach}}} & AID-R & \textbf{75.41\%} & 73.56\% & 69.76\% \\
        &   AID-D & 70.46\% & \textbf{74.42\%} & 67.42\% \\
        &   AID-I & 69.95\% & 69.88\% & \textbf{73.12\%} \\
    \bottomrule
  \end{tabular}
  \caption{\textbf{Cross Model Attack Identification.} AID-R, AID-D, and AID-I refer to the subsets of AID containing perturbations corresponding to the target models ResNet50\cite{resnet}, DenseNet121\cite{dense} (abbreviated as DenNet121) and InceptionV3\cite{incept} respectively.}
  \label{tab:Cross-model}
  \vspace{-3mm}
\end{table}



\vspace{-1mm}
\subsection{Analysis and Discussion}
\vspace{-1mm}
{\bf Confusion Matrix:}
We analyze class wise scores and the confusion matrix of the predictions from the proposed approach in Fig \ref{fig:conf}. From the confusion matrix, we observe the common trend of relatively high scores for all decision based attacks except for boundary attack. With scores close to 1, these attacks have distinctive patterns which are being easily identified by the signature extractor. Boundary attack do not always have specific patterns because of the way they are generated. Boundary attack performs a random walk on the decision boundary minimizing the amount of perturbation. Similarly, universal attacks generate discernible patterns making it easier for detection. Major confusion occurs in the gradient based attacks among NewtonFool, DeepFool and CW attack. These attacks being highly powerful, are targeted on generated nearly imperceptible perturbations specific to the input image, making it difficult for the method to identify and distinguish.

{\bf Signature Extraction:}
Table \ref{tab:abl1} investigates the performance of the signature extractor under various settings. Standard GLOF achieves higher PSNR and SSIM scores over GLOF-C and GLOF-A indicating that global and local connectivity used in conjunction help in better reconstruction. We also report the variation in reconstruction scores when the number of heads $m$ in multi head attention are increased. GLoF modules with 12 heads achieves the highest scores of \textbf{31.55} PSNR and \textbf{0.89} SSIM. \par
{\bf Number of GLOF modules:} 
We analyze the performance of the network by varying the number of GLOF modules. Signature Extractor with as low as a single GLOF module achieves 79.20\% (+6\% over baseline) thus indicating its effectiveness. Employing 5 GLoF modules yields the best accuracy of 80.14\%. \par 
{\bf Cross model attack identification:} 
We analyze the performance of our network on cross model attack identification. AID consists of attacks generated by targeting 3 different networks. For this experiment, we split AID into three subsets containing perturbations related to the corresponding target model. AID-R, AID-D, AID-I refer to subsets of AID containing perturbations corresponding to ResNet50\cite{resnet}, DenseNet121\cite{dense} and InceptionV3\cite{incept} as target networks. Each subset is further split into train-test sets. Table \ref{tab:Cross-model} details the results on cross model attack identification of several baselines compared against our technique. In general, we observe that the networks perform well when trained and tested on the same subsets of AID. The proposed technique performs better in all cases compared to other baselines. This experiment suggests that perturbations from different target models also contain similar traces that can be leveraged to profile the attacker. \par

\begin{figure}[t!]
  \centering
         \includegraphics[width=\linewidth]{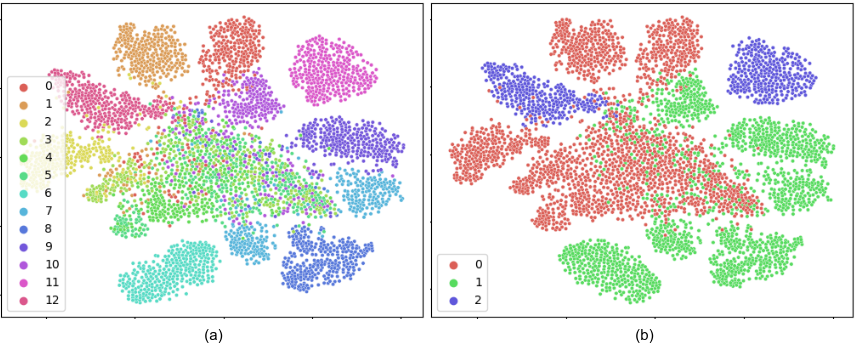}

\caption{\textbf{Visualizations of features learned by the attack classifier.} (a) t-SNE for specific attack categories. Labels are according to Table \ref{tab:attacks1} (b) t-SNE for attack families. Labels \{0,1,2\} refer to \{gradient, decision, universal\} attacks.
}
\label{fig:tsne}
\vspace{-3mm}
\end{figure}

\textbf{Success rate vs. Identifiability}: While the stronger attacks like PGD have a 100\% fooling rate, the weaker black box attacks have a success rate of atleast 65\% for the samples considered in AID. We also study the indentifiability vs success rate for the FGSM class and find that our technique achieves 74.9\% accuracy for an epsilon as low as 2 and 94.5\% for an epsilon of value 16. We observe an upward increasing trend as the epsilon increases indicating an increasing level of perceptibility of the patterns.

{\bf Identifying unseen attacks:} With the increasing threat to neural networks, it is likely for the PRAT problem to encounter novel/unseen attacks. To experiment the effectiveness of the proposed network we devise an experiment which includes identifying the toolchain family of an unseen attack. For this, we split AID into two different sets containing mutually exclusive attack categories. We retrain the overall pipeline and test it on the unseen classes which achieves an accuracy of 57.2\%. We extend our approach to register novel attacks with minimal training set using toolchain indexing(discussed in supplementary). Identifying open set novel attacks under PRAT scenario remains challenging due to the fact that the unseen perturbations are nearly imperceptible and are difficult to distinguish.

{\bf Feature visualization:} We study the separability of extracted features by analyzing the t-SNE plots of a set of features extracted from the penultimate layer of the attack classifier. Fig.\ref{fig:tsne} shows the three toolchain families forming separate clusters. Due to their `universality' constraint, universal perturbations form a clear cluster and are easily distinguishable. While gradient based attacks share similar techniques, decision based attacks have distinctive approaches based on the decision of the network. Hence we observe the overlap between gradient and decision based attacks.  Fig.\ref{fig:tsne} shows the t-SNE plots over specific classes. Boundary Attack has the maximum overlap with other attacks. In gradient based attacks, DeepFool, NewtonFool and CW attacks overlap with each other indicating that they generate similar patterns thus making it difficult to distinguish them.  

\begin{figure}[t!]
  \centering
         \includegraphics[width=0.9\linewidth]{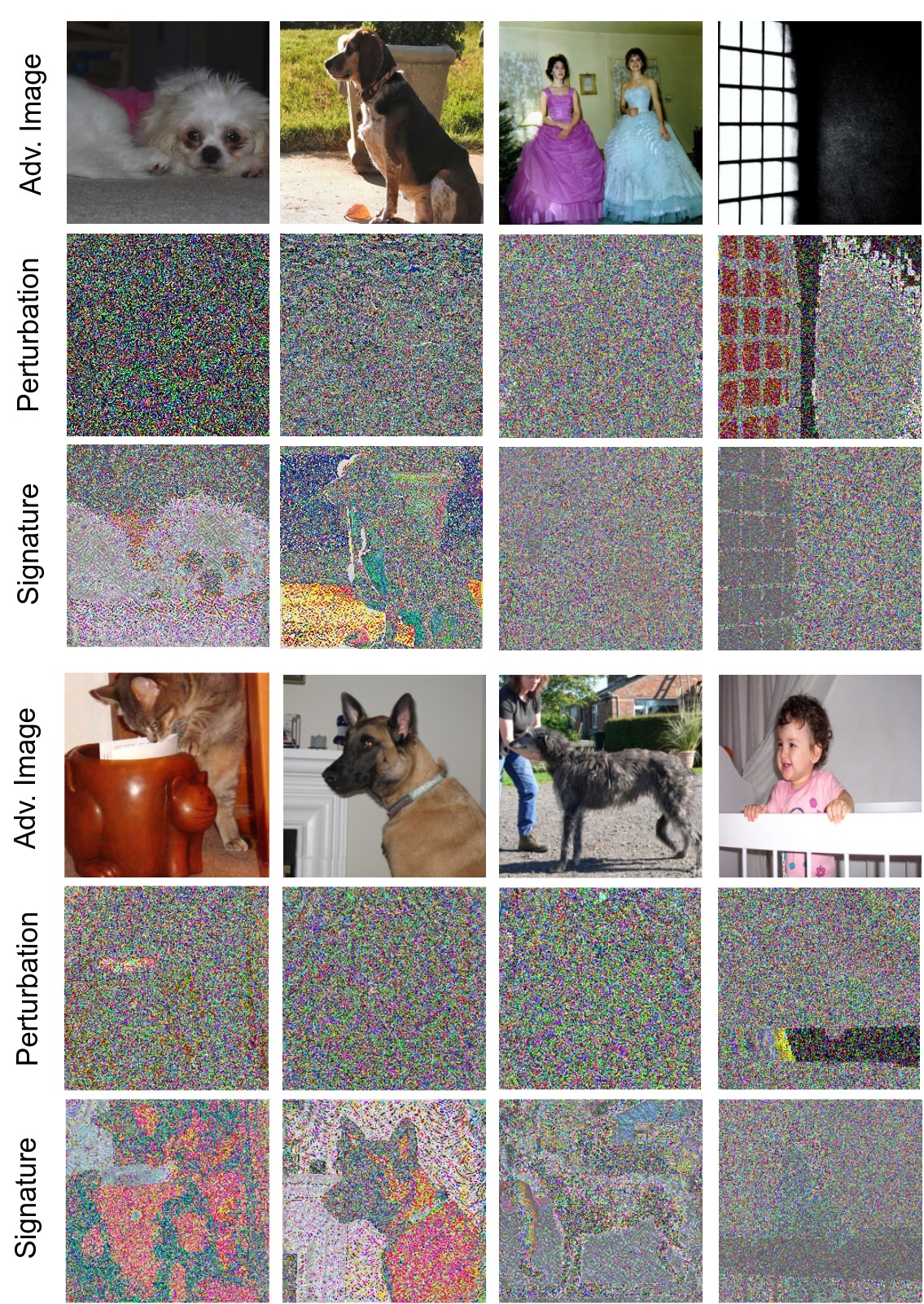}
\caption{Adversarial images, their perturbations (normalized) and the corresponding signature(normalized) extracted by the proposed approach.}
\label{fig:recons}
\vspace{-10pt}
\end{figure}

{\bf Reconstructions:}
Fig \ref{fig:recons}. depicts the adversarial images, corresponding perturbations and the signatures extracted by the our method. In general, the extracted signatures have patterns highlighting the object from the image. This is due to the fact that extracting these nearly imperceptible perturbations accurately is always challenging. These patterns along with the patterns pertaining to the attacker help in training the attack classifier to identify the attacker.

\vspace{-2mm}
\section{Conclusion}
\vspace{-1mm}
We presented a new perspective on adversarial attacks indicating the presence of peculiar patterns in the perturbations that hint back to the attacker. We formulate the PRAT problem - given the adversarial input, profile the attack signature to identify the attack used to generate the sample. We develop Adversarial Identification Dataset and compare several baseline techniques on the proposed dataset. Targeting PRAT, we propose a framework that combines CNN's capability to capture local features and Transformer's ability to encode global attention to generate signatures containing attack-specific patterns, which are used by an attack classifier to identify the attack. Extensive experiments showcase the efficacy of our framework and support the credibility of the proposed PRAT problem.


{\small
\bibliographystyle{ieee_fullname}
\bibliography{egbib}

\begin{thebibliography}{10}\itemsep=-1pt

\bibitem{akhtar2018threat}
Naveed Akhtar and Ajmal Mian.
\newblock Threat of adversarial attacks on deep learning in computer vision: A survey.
\newblock {\em Ieee Access}, 6:14410--14430, 2018.

\bibitem{akhtar2021advances}
Naveed Akhtar, Ajmal Mian, Navid Kardan, and Shah Mubarak.
\newblock Advances in adversarial attacks and defenses in computer vision: A survey.
\newblock {\em arXiv preprint arXiv:2108.00401v2}, 2021.

\bibitem{ACFH2020square}
Maksym Andriushchenko, Francesco Croce, Nicolas Flammarion, and Matthias Hein.
\newblock Square attack: a query-efficient black-box adversarial attack via random search.
\newblock 2020.

\bibitem{yang}
Yang Bai, Yuyuan Zeng, Yong Jiang, Yisen Wang, Shu-Tao Xia, and Weiwei Guo.
\newblock Improving query efficiency of black-box adversarial attack.
\newblock In Andrea Vedaldi, Horst Bischof, Thomas Brox, and Jan-Michael Frahm, editors, {\em Computer Vision -- ECCV 2020}, pages 101--116, Cham, 2020. Springer International Publishing.

\bibitem{brendel2018decisionbased}
Wieland Brendel, Jonas Rauber, and Matthias Bethge.
\newblock Decision-based adversarial attacks: Reliable attacks against black-box machine learning models.
\newblock In {\em International Conference on Learning Representations}, 2018.

\bibitem{carlini2017towards}
Nicholas Carlini and David Wagner.
\newblock Towards evaluating the robustness of neural networks.
\newblock In {\em 2017 ieee symposium on security and privacy (sp)}, pages 39--57. IEEE, 2017.

\bibitem{carlini}
Nicholas Carlini and David Wagner.
\newblock Towards evaluating the robustness of neural networks.
\newblock In {\em 2017 IEEE Symposium on Security and Privacy (SP)}, pages 39--57, 2017.

\bibitem{chakraborty2018adversarial}
Anirban Chakraborty, Manaar Alam, Vishal Dey, Anupam Chattopadhyay, and Debdeep Mukhopadhyay.
\newblock Adversarial attacks and defences: A survey.
\newblock {\em arXiv preprint arXiv:1810.00069}, 2018.

\bibitem{chakraborty2021survey}
Anirban Chakraborty, Manaar Alam, Vishal Dey, Anupam Chattopadhyay, and Debdeep Mukhopadhyay.
\newblock A survey on adversarial attacks and defences.
\newblock {\em CAAI Transactions on Intelligence Technology}, 6(1):25--45, 2021.

\bibitem{Chen2020BoostingDB}
Weilun Chen, Zhaoxiang Zhang, Xiaolin Hu, and Baoyuan Wu.
\newblock Boosting decision-based black-box adversarial attacks with random sign flip.
\newblock In {\em ECCV}, 2020.

\bibitem{Croce2020Provable}
Francesco Croce and Matthias Hein.
\newblock Provable robustness against all adversarial $l_p$-perturbations for $p\geq 1$.
\newblock In {\em International Conference on Learning Representations}, 2020.

\bibitem{deng2021libre}
Zhijie Deng, Xiao Yang, Shizhen Xu, Hang Su, and Jun Zhu.
\newblock Libre: A practical bayesian approach to adversarial detection.
\newblock In {\em Proceedings of the IEEE/CVF Conference on Computer Vision and Pattern Recognition}, pages 972--982, 2021.

\bibitem{Devlin2019BERTPO}
Jacob Devlin, Ming-Wei Chang, Kenton Lee, and Kristina Toutanova.
\newblock Bert: Pre-training of deep bidirectional transformers for language understanding.
\newblock In {\em NAACL}, 2019.

\bibitem{dosovitskiy2020vit}
Alexey Dosovitskiy, Lucas Beyer, Alexander Kolesnikov, Dirk Weissenborn, Xiaohua Zhai, Thomas Unterthiner, Mostafa Dehghani, Matthias Minderer, Georg Heigold, Sylvain Gelly, Jakob Uszkoreit, and Neil Houlsby.
\newblock An image is worth 16x16 words: Transformers for image recognition at scale.
\newblock {\em ICLR}, 2021.

\bibitem{du2019query}
Jiawei Du, Hu Zhang, Joey~Tianyi Zhou, Yi Yang, and Jiashi Feng.
\newblock Query-efficient meta attack to deep neural networks.
\newblock {\em arXiv preprint arXiv:1906.02398}, 2019.

\bibitem{goodfellow2014explaining}
Ian Goodfellow, Jonathon Shlens, and Christian Szegedy.
\newblock Explaining and harnessing adversarial examples.
\newblock {\em arXiv preprint arXiv:1412.6572}, 2014.

\bibitem{universal}
J. Hayes and G. Danezis.
\newblock Learning universal adversarial perturbations with generative models.
\newblock In {\em 2018 IEEE Security and Privacy Workshops (SPW)}, pages 43--49, Los Alamitos, CA, USA, may 2018. IEEE Computer Society.

\bibitem{resnet}
Kaiming He, Xiangyu Zhang, Shaoqing Ren, and Jian Sun.
\newblock Deep residual learning for image recognition.
\newblock In {\em Proceedings of the IEEE conference on computer vision and pattern recognition}, pages 770--778, 2016.

\bibitem{dense}
Gao Huang, Zhuang Liu, Laurens Van Der~Maaten, and Kilian~Q Weinberger.
\newblock Densely connected convolutional networks.
\newblock In {\em Proceedings of the IEEE conference on computer vision and pattern recognition}, pages 4700--4708, 2017.

\bibitem{Huang2020Black-Box}
Zhichao Huang and Tong Zhang.
\newblock Black-box adversarial attack with transferable model-based embedding.
\newblock In {\em International Conference on Learning Representations}, 2020.

\bibitem{newton}
Uyeong Jang, Xi Wu, and Somesh Jha.
\newblock Objective metrics and gradient descent algorithms for adversarial examples in machine learning.
\newblock In {\em Proceedings of the 33rd Annual Computer Security Applications Conference}, ACSAC 2017, page 262–277, New York, NY, USA, 2017. Association for Computing Machinery.

\bibitem{kannan2018adversarial}
Harini Kannan, Alexey Kurakin, and Ian Goodfellow.
\newblock Adversarial logit pairing.
\newblock {\em arXiv preprint arXiv:1803.06373}, 2018.

\bibitem{katz2017reluplex}
Guy Katz, Clark Barrett, David~L Dill, Kyle Julian, and Mykel~J Kochenderfer.
\newblock Reluplex: An efficient smt solver for verifying deep neural networks.
\newblock In {\em International Conference on Computer Aided Verification}, pages 97--117. Springer, 2017.

\bibitem{krizhevsky2012imagenet}
Alex Krizhevsky, Ilya Sutskever, and Geoffrey~E Hinton.
\newblock Imagenet classification with deep convolutional neural networks.
\newblock {\em Advances in neural information processing systems}, 25:1097--1105, 2012.

\bibitem{kurakin2016adversarial}
Alexey Kurakin, Ian Goodfellow, Samy Bengio, et~al.
\newblock Adversarial examples in the physical world, 2016.

\bibitem{lecun1995convolutional}
Yann LeCun, Yoshua Bengio, et~al.
\newblock Convolutional networks for images, speech, and time series.
\newblock {\em The handbook of brain theory and neural networks}, 3361(10):1995, 1995.

\bibitem{li2020qeba}
Huichen Li, Xiaojun Xu, Xiaolu Zhang, Shuang Yang, and Bo Li.
\newblock Qeba: Query-efficient boundary-based blackbox attack.
\newblock In {\em Proceedings of the IEEE/CVF Conference on Computer Vision and Pattern Recognition}, pages 1221--1230, 2020.

\bibitem{DBLP:conf/eccv/LiZPRSKSC20}
Shasha Li, Shitong Zhu, Sudipta Paul, Amit~K. Roy{-}Chowdhury, Chengyu Song, Srikanth~V. Krishnamurthy, Ananthram Swami, and Kevin~S. Chan.
\newblock Connecting the dots: Detecting adversarial perturbations using context inconsistency.
\newblock In Andrea Vedaldi, Horst Bischof, Thomas Brox, and Jan{-}Michael Frahm, editors, {\em Computer Vision - {ECCV} 2020 - 16th European Conference, Glasgow, UK, August 23-28, 2020, Proceedings, Part {XXIII}}, volume 12368 of {\em Lecture Notes in Computer Science}, pages 396--413. Springer, 2020.

\bibitem{li2021localvit}
Yawei Li, Kai Zhang, Jiezhang Cao, Radu Timofte, and Luc Van~Gool.
\newblock Localvit: Bringing locality to vision transformers.
\newblock {\em arXiv preprint arXiv:2104.05707}, 2021.

\bibitem{ma2021simulator}
Chen Ma, Li Chen, and Jun-Hai Yong.
\newblock Simulating unknown target models for query-efficient black-box attacks.
\newblock In {\em Proceedings of the IEEE/CVF Conference on Computer Vision and Pattern Recognition (CVPR)}, pages 11835--11844, June 2021.

\bibitem{madry2017towards}
Aleksander Madry, Aleksandar Makelov, Ludwig Schmidt, Dimitris Tsipras, and Adrian Vladu.
\newblock Towards deep learning models resistant to adversarial attacks.
\newblock {\em arXiv preprint arXiv:1706.06083}, 2017.

\bibitem{deep}
S. Moosavi-Dezfooli, A. Fawzi, and P. Frossard.
\newblock Deepfool: A simple and accurate method to fool deep neural networks.
\newblock In {\em 2016 IEEE Conference on Computer Vision and Pattern Recognition (CVPR)}, pages 2574--2582, Los Alamitos, CA, USA, jun 2016. IEEE Computer Society.

\bibitem{Moosavi-Dezfooli_2017_CVPR}
Seyed-Mohsen Moosavi-Dezfooli, Alhussein Fawzi, Omar Fawzi, and Pascal Frossard.
\newblock Universal adversarial perturbations.
\newblock In {\em Proceedings of the IEEE Conference on Computer Vision and Pattern Recognition (CVPR)}, July 2017.

\bibitem{moosavi2016deepfool}
Seyed-Mohsen Moosavi-Dezfooli, Alhussein Fawzi, and Pascal Frossard.
\newblock Deepfool: a simple and accurate method to fool deep neural networks.
\newblock In {\em Proceedings of the IEEE conference on computer vision and pattern recognition}, pages 2574--2582, 2016.

\bibitem{papernot2016limitations}
Nicolas Papernot, Patrick McDaniel, Somesh Jha, Matt Fredrikson, Z~Berkay Celik, and Ananthram Swami.
\newblock The limitations of deep learning in adversarial settings.
\newblock In {\em 2016 IEEE European symposium on security and privacy (EuroS\&P)}, pages 372--387. IEEE, 2016.

\bibitem{Qin2020Detecting}
Yao Qin, Nicholas Frosst, Sara Sabour, Colin Raffel, Garrison Cottrell, and Geoffrey Hinton.
\newblock Detecting and diagnosing adversarial images with class-conditional capsule reconstructions.
\newblock In {\em International Conference on Learning Representations}, 2020.

\bibitem{rahmati2020geoda}
Ali Rahmati, Seyed-Mohsen Moosavi-Dezfooli, Pascal Frossard, and Huaiyu Dai.
\newblock Geoda: a geometric framework for black-box adversarial attacks.
\newblock In {\em Proceedings of the IEEE/CVF Conference on Computer Vision and Pattern Recognition}, pages 8446--8455, 2020.

\bibitem{rauber2017foolbox}
Jonas Rauber, Wieland Brendel, and Matthias Bethge.
\newblock Foolbox: A python toolbox to benchmark the robustness of machine learning models.
\newblock In {\em Reliable Machine Learning in the Wild Workshop, 34th International Conference on Machine Learning}, 2017.

\bibitem{ILSVRC15}
Olga Russakovsky, Jia Deng, Hao Su, Jonathan Krause, Sanjeev Satheesh, Sean Ma, Zhiheng Huang, Andrej Karpathy, Aditya Khosla, Michael Bernstein, Alexander~C. Berg, and Li Fei-Fei.
\newblock {ImageNet Large Scale Visual Recognition Challenge}.
\newblock {\em International Journal of Computer Vision (IJCV)}, 115(3):211--252, 2015.

\bibitem{shi2020polishing}
Yucheng Shi, Yahong Han, and Qi Tian.
\newblock Polishing decision-based adversarial noise with a customized sampling.
\newblock In {\em Proceedings of the IEEE/CVF Conference on Computer Vision and Pattern Recognition}, pages 1030--1038, 2020.

\bibitem{silver2017mastering}
David Silver, Julian Schrittwieser, Karen Simonyan, Ioannis Antonoglou, Aja Huang, Arthur Guez, Thomas Hubert, Lucas Baker, Matthew Lai, Adrian Bolton, et~al.
\newblock Mastering the game of go without human knowledge.
\newblock {\em nature}, 550(7676):354--359, 2017.

\bibitem{incept}
Christian Szegedy, Vincent Vanhoucke, Sergey Ioffe, Jon Shlens, and Zbigniew Wojna.
\newblock Rethinking the inception architecture for computer vision.
\newblock In {\em Proceedings of the IEEE conference on computer vision and pattern recognition}, pages 2818--2826, 2016.

\bibitem{szegedy2013intriguing}
Christian Szegedy, Wojciech Zaremba, Ilya Sutskever, Joan Bruna, Dumitru Erhan, Ian Goodfellow, and Rob Fergus.
\newblock Intriguing properties of neural networks.
\newblock {\em arXiv preprint arXiv:1312.6199}, 2013.

\bibitem{tjeng2018evaluating}
Vincent Tjeng, Kai~Y. Xiao, and Russ Tedrake.
\newblock Evaluating robustness of neural networks with mixed integer programming.
\newblock In {\em International Conference on Learning Representations}, 2019.

\bibitem{tramer2018ensemble}
Florian Tramèr, Alexey Kurakin, Nicolas Papernot, Ian Goodfellow, Dan Boneh, and Patrick McDaniel.
\newblock Ensemble adversarial training: Attacks and defenses.
\newblock In {\em International Conference on Learning Representations}, 2018.

\bibitem{vaswani2017attention}
Ashish Vaswani, Noam Shazeer, Niki Parmar, Jakob Uszkoreit, Llion Jones, Aidan~N Gomez, {\L}ukasz Kaiser, and Illia Polosukhin.
\newblock Attention is all you need.
\newblock In {\em Advances in neural information processing systems}, pages 5998--6008, 2017.

\bibitem{hanruiwang2020hat}
Hanrui Wang, Zhanghao Wu, Zhijian Liu, Han Cai, Ligeng Zhu, Chuang Gan, and Song Han.
\newblock Hat: Hardware-aware transformers for efficient natural language processing.
\newblock In {\em Annual Conference of the Association for Computational Linguistics}, 2020.

\bibitem{zhang2019theoretically}
Hongyang Zhang, Yaodong Yu, Jiantao Jiao, Eric Xing, Laurent El~Ghaoui, and Michael Jordan.
\newblock Theoretically principled trade-off between robustness and accuracy.
\newblock In {\em International Conference on Machine Learning}, pages 7472--7482. PMLR, 2019.

\end{thebibliography}
}

\end{document}